# An Explainable Deep Learning-based Prognostic Model for Rotating Machinery

Namkyoung Lee, Michael H. Azarian, *Member, IEEE,* and Michael Pecht, *Fellow, IEEE*

*Abstract—* **This paper develops an explainable deep learning model that estimates the remaining useful lives of rotating machinery. The model extracts high-level features from Fourier transform using an autoencoder. The features are used as input to a feedforward neural network to estimate the remaining useful lives. The paper explains the model's behavior by analyzing the composition of the features and the relationships between the features and the estimation results. In order to make the model explainable, the paper introduces octave-band filtering. The filtering reduces the input size of the autoencoder and simplifies the model. A case study demonstrates the methods to explain the model. The study also shows that the octave-band filtering in the model imitates the functionality of low-level convolutional layers. This result supports the validity of using the filtering to reduce the depth of the model.**

## I. INTRODUCTION

Deep learning has successfully been used to estimate the remaining useful lives (RULs) of rotating machinery regardless of the types of neural network structures [1-9]. In general, this success is attributed to high flexibility in modeling features for better RUL estimation [10]. However, deep learning makes the estimation process opaque. The complex connections among multiple neural layers deter our understanding of how deep learning estimates RULs. This problem is linked to the credibility of deep learning. For example, a deep learning model may learn spurious correlations while training due to an inadequate amount of training data. If humans cannot detect the spurious correlations and correct them, the model may give inaccurate RUL estimates in real-world applications.

Many methods have been developed to explain the decision logic of deep learning models. One way to explain the behavior of a deep learning model is to measure the inputs' importance factors that help determine the model's output. Methods such as layer-wise relevance propagation (LRP) [11], local interpretable model-agnostic explanations [12], and Shapley additive explanations [13] reveal why a model made the decision with the given inputs. These methods have been applied to convolutional neural network (CNN)-based diagnostics for rotating machinery. Saeki et al. [6] and Grezmak et al. [7] used gradient-weighted class activation mapping and LRP to analyze attributions of frequency bands in spectrograms to diagnosis results. Although their work explained which inputs affect the results most, they were not able to explain the high-level features of their trained deep learning models.

Visualizing the features of the developed neural networks is another approach to understand deep learning models. Jia et al. [8] visualized the features of a CNN by feeding inputs to the network that maximize the activation functions of the convolutional layers. Lei et al. [9] correlated the functionality of their sparse filtering to that of Gabor filters. They showed the physical representation of the features of the sparse filtering by modeling the weights of the filtering to a Gabor filter. This approach gave insights about how deep learning models interpret the inputs, especially for deep learning-based diagnostics. However, deep learning-based prognostics require additional explanation to connect the relationship between the features and the results of the prognostics.

This paper develops a method that explains the relationship by reducing the size of a neural network to the level where a human can understand the network. In general, a neural network that has a hierarchical structure learns different types of features depending on the location of the layer. For example, LeCun et al. [14] showed that a CNN could learn the edges and lines of images from the first few layers, and the layers at the top of the network were composed of patterns using the edges and lines.

One method to reduce the size of a neural network is to modify the input of the network by utilizing domain knowledge. The schematic in Fig. 1 shows hierarchical neural networks that were trained for the same purpose. If the hand-crafted features on the right replicate the outputs of the low-level layers of the neural network on the left, the network on the right can perform the same function while reducing the depth of the network by directly receiving the hand-crafted features as inputs.

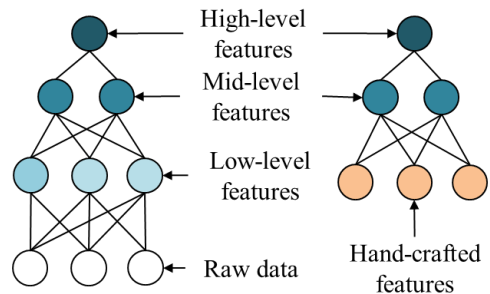

Fig. 1. A schematic of hierarchical neural networks with annotations about the characteristics of features for each level. The low-level features on the left neural network can be replaced by hand-crafted features to reduce the depth of the neural network.

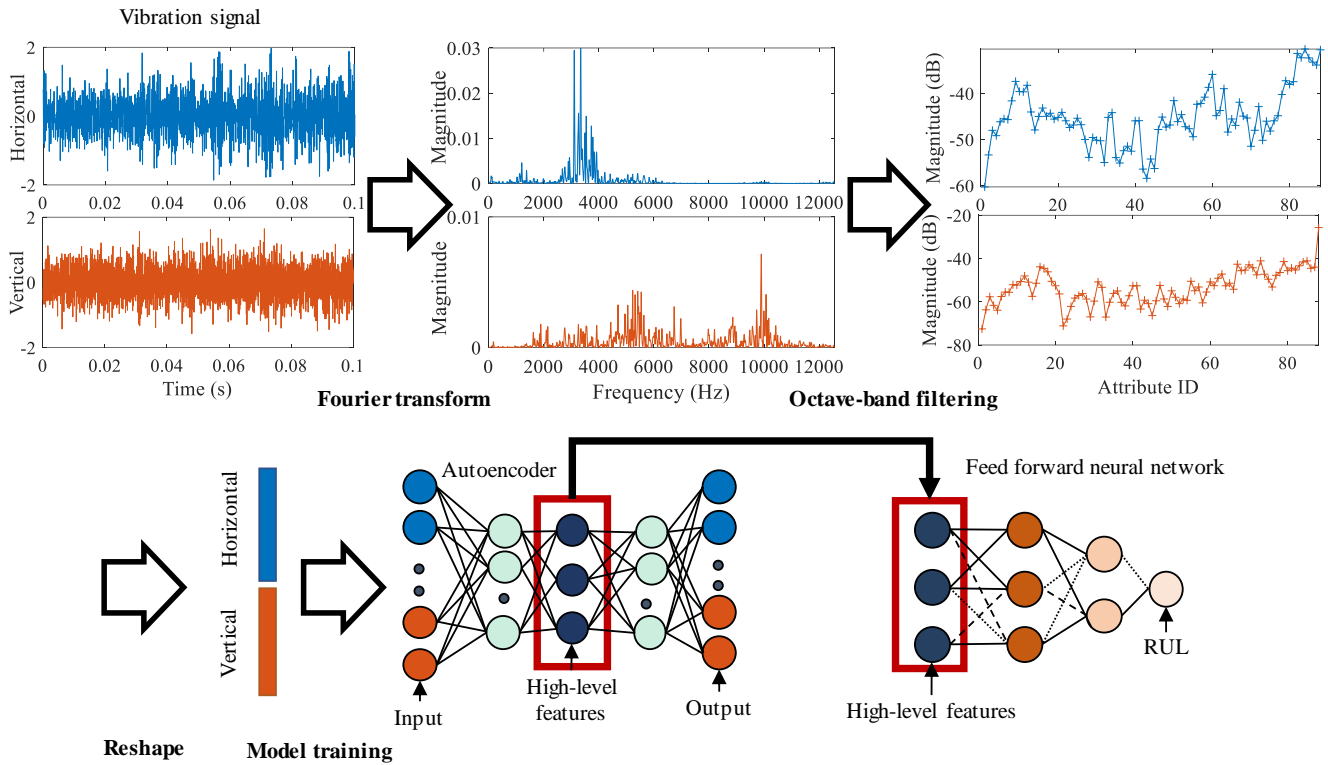

Fig. 2. An overall process to build an explainable deep learning-based prognostics model. The model receives two dimensional accelerations and outputs the remaining useful life of a system.

Previous studies [6, 7] have explained that deep learning models that diagnose rotating machinery find characteristic frequencies for bearing faults from vibrations nearby bearings. This paper hypothesizes that the low-level layers of the models find local correlations among frequency components that are related to faults of the machinery. Based on this hypothesis, the paper develops a deep learning-based prognostics model for rotating machinery that utilizes octave-band filtering to replace the functionality of the low-level layers. Octave-band filtering extracts features of the characteristic frequencies while eliminating redundancy among them.

In order to interpret the network, the paper uses an autoencoder (AE) [15] and a feedforward neural network (FFNN). The AE generates high-level features from the results of octave-band filtering, and the FFNN estimates RULs of rotating machinery. As the size of the network is compact, the composition of the high-level features can be explained by visualizing the features. In addition, the attributions of the features to the RUL estimations can be calculated. Through these processes, humans can inspect and correct the trained neural networks and improve the performance of the networks. The paper provides a case study to explain how to interpret the developed model.

The rest of the paper is organized as follows. Section II introduces the overall process to build an explainable deep learning-based prognostics model. After the introduction, octave-band filtering is elaborated as a dimensionality reduction method, especially for rotating machinery signals. The methods to inspect the model are also explained. Section III presents a case study to show how the developed model is explainable. Section IV discusses the results of the case study. Section V concludes the paperwork.

## II. METHOD

The overall process to estimate RULs of rotating machinery using a deep learning model is depicted in Fig. 2. The model receives signals from accelerometers and generates power spectra from the signals through Fourier transform. In order to reduce the size of the model's neural network, octave-band filtering compresses the power spectra. The outputs of the octave-band filtering at the same timeframe are concatenated to be fed to an autoencoder (AE) for training. After the training, high-level features of the AE are extracted from the bottleneck of the AE by inputting the training data. The features are used as inputs for a feedforward neural network (FFNN), and the FFNN outputs the estimated RUL of the machinery. The FFNN was trained in a supervised manner. The RULs that are associated with the training inputs were calculated backward from the point where the machinery is failed.

### A. Octave-band Filtering

In general, most of the frequency components from the vibrations of rotating machinery are inactive and do not contain useful information for diagnosis and prognosis. They also have redundant information. For example, characteristic frequencies of rolling element bearings, such as the ball pass frequency of the outer race, often have sidebands and harmonics, which deliver the same information. To handle this problem, octave bands filtered the power spectra after the Fourier transform.

Octave-band Filtering is one of the methods for analyzing vibration in the frequency domain, which is defined in ANSI standard S1.11-2004 [16]. The method bins a power spectrum within certain frequency bands that are logarithmically scaled.

The relationship between adjacent band edges of $n$th octave bands is defined as follows:

$$\frac{f_{i+1}}{f_i} = 2^{\frac{1}{n}}, i = 1,2,3, \dots, N \tag{1}$$

where $i$ represents the order of frequency band edges, and $n$ is the number of bands in one octave. As $n$ increases, the resolution of octave-band filtering increases. Based on the relationship, the locations of the edges are determined by defining the first band edge $f_1$ and the number of bands $N$.

Octave-band filtering sums coefficients of a power spectrum between the octave frequency band edges. Since the model uses discrete power spectral densities as the input of the octave-band filtering, the filtering process proceeds as follows:

$$F_i = \frac{1}{b-a}\sum_{k=a}^{b} psd(k) \tag{2}$$
$$a = \min_i f(i) > f_i, b = \min_i f(i) > f_{i+1}$$

where $psd(k)$ and $f(i)$ represent the $k$th power spectral density and the $i$th upper edge frequency of the power spectral density.

Octave band filters usually stack the power spectral densities of the characteristic frequencies and their sidebands together, which reduces redundancy. They also generate features that are related to resonance frequencies of rotating machinery elements. Take a rolling element bearing's power spectra as an example. The power spectral densities above around 3 kHz are related to the natural frequencies of the bearing element's materials [17]. The locations of the natural frequencies can smear in the observed power spectra due to operating conditions and the lack of spectra resolution. Octave-band filtering accumulates a wide range of power spectral densities that are related to the resonance frequencies. Therefore, the changes in the power spectral densities that are related to the resonance frequencies can be tracked.

This paper uses a modified octave filter to reduce the redundancy in octave bands in the low region. These bands are usually narrower due to logarithmic scaling. Therefore, the modified filter may not eliminate redundancy among the fault frequencies at low-frequency bands and their sidebands. To increase the band width, the modified filter uses octave bands that have a constant absolute band width in the low region. The edges of the octave bands can be expressed using formulae as follows:

$$f_i = m(i-1), i = 1,2,3, \dots, N$$
$$f_j = 2^{\frac{j+k}{n}}, j = 1,2,3, \dots, M \tag{3}$$
$$k = \left\lceil n \cdot \log_2 \frac{m}{2^{1/n}-1} \right\rceil, N = \left\lfloor \frac{1}{m} \cdot 2^{\frac{k+1}{n}} + 1 \right\rfloor$$

where $f_i$ and $f_j$ are the band edges of different types of octave bands. The modified octave bands are constructed by concatenating the two band edges in order. The first $N$-1 octave bands have constant absolute band width of $m$ Hz, and the rest of the octave bands have constant percentage band width that follows (1). The octave band index $N$ where the constant percentage band width is applied first is defined by the point where the band width of octave filters in (1) having $f_1 = 1$ exceeds the constant band width $m$.

## B. High-level Feature Explanation

The developed model uses an AE to generate high-level features from the results of octave-band filtering. An AE is a type of neural network consisting of an encoder and a decoder. An encoder compresses inputs into high-level features, and a decoder restores the inputs using the features. The characteristics of high-level features can be explained by utilizing the characteristics of an AE. Overall, the values of the high-level features are stable when inputting the vibration signals of a healthy rotating machine to the model because the signals are stable. As the machine degrades, the signals may exhibit unusual patterns in power spectra, which leads to changes in the features. By correlating the changes in the inputs with the outputs of an encoder, the composition of the features can be explained. However, this method may not segregate the characteristics of the features because more than one feature value can change simultaneously as the machine degrades.

Another method to explain the high-level features is reversing the process using a decoder. As a decoder recovers the input of an encoder using the high-level features, the characteristics of each feature can be observed by changing one feature value at a time. Compared to the previous method, humans can only infer the characteristics of the features indirectly because the decoding process is not an exact inverse of the encoding process. The explanation becomes uncertain when the trained decoder is overfitted to the training data. The overfitted decoder generates erroneous outputs with small variations of the inputs. As this method manually feeds modified inputs to a decoder, the outputs of the decoder can also be erroneous. Therefore, the characteristics of the high-level features of an AE can only be explained if these two methods are used together.

## C. Feature Importance Evaluation

The FFNN in the model estimates RUL based on the high-level features of the AE. In order to inspect and correct the decision logic of the FFNN, the contribution of the features to the RUL estimate can be evaluated. Olden et al. [18] conducted a comparative study to evaluate algorithms that evaluate feature importance for an FFNN. Among them, the connection weight approach [19] provided the best accuracy in quantifying variable importance in the study.

The connection weight algorithm measures feature importance of one-layer neural network by summing the connection weights between input and hidden nodes and then weighting the results by the connection weights between hidden and output nodes. The algorithm can be expressed as follows:

$$FI_i = \sum_{j=1}^{M} w_j^2 w_{i,j}^1, j = 1,2, \dots, M \tag{4}$$

where $i$ and $j$ represent the index of inputs and hidden neurons. $w^1$, $w^2$, and $FI$ represent input-hidden and hidden-output connection weights and their relative importance. For example, the feature importance of the first input $FI_1$ is the summation of weights that are connected to the first neuron $w_{1,j}^1$, multiplied by the weights $w_j^2$ that are associated with the connected node.

Although Olden et al. applied the algorithm to a single-layer FFNN, the algorithm can also be applied to a multi-layer FFNN by conducting the algorithm recursively from the output node to input nodes. The generic formula to evaluate the feature importance of a $l$th layer FFNN can be defined as follows:

$$FI_i^l = \sum_{j=1}^{N_{l+1}} FI_j^{l+1} w_{i,j}^l,$$
$$j = 1,2, \dots, M_l, l = 1,2, \dots, L$$
$$FI_i = FI_i^1, FI_j^{L+1} = w_j^L$$

(5)

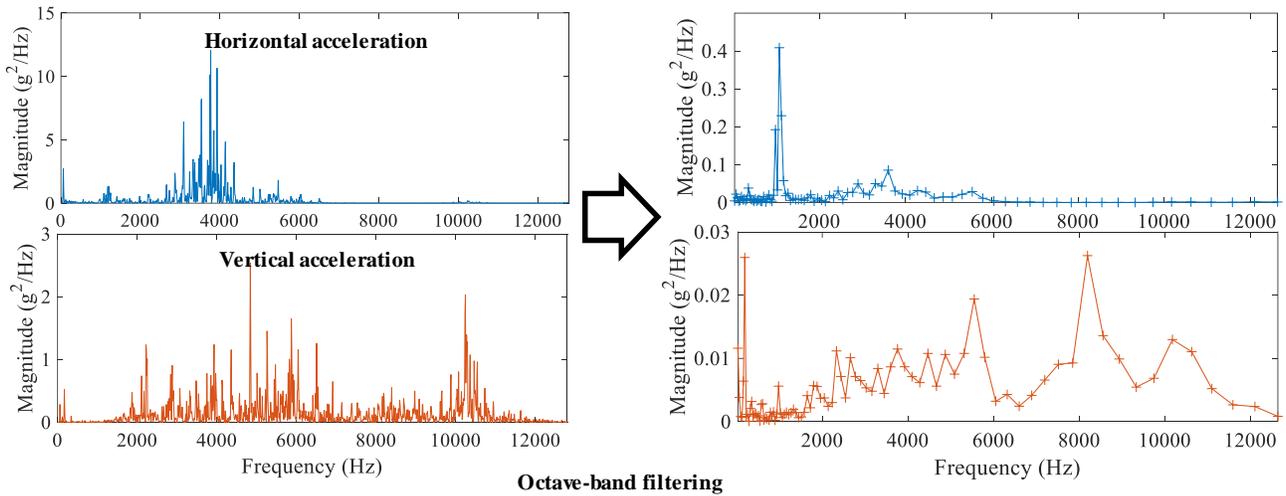

Fig. 3. Octave-band filtering for a bearing's spectral data. The octave-band filtering reduces the dimensionality of the original data while reserving the overall peaks of the data.

Compared to (4), (5) uses a generic weight $FI_j^{l+1}$ that represents the feature importance of the $j$th node at the $l+1$th layer. The calculation process starts by obtaining the feature importance of the last hidden layer $FI_i^L$ and ends by obtaining the feature importance of the input node $FI_i^1$, which is the feature importance of the FFNN.

The obtained feature importance can be regarded as the correlation between the input and the output of an FFNN. As the feature importance of the input increases, the effect of the input change on the output also increases. If feature importance is negative, the output of an FFNN decreases as the input increases. However, interpreting the feature importance becomes difficult as the number of layers of an FFNN increases because the algorithm omits the nonlinearity of activation functions in the FFNN. The existence of activation functions increases the complexity of interpreting the relationship between the input and output of an FFNN. Therefore, the application of this algorithm should be limited to an FFNN that has few neural layers.

## III. CASE STUDY

The IEEE PHM 2012 data challenge bearing dataset [20] was used to validate the developed method. The dataset consists of data from 6 run-to-failure tests under three different loading conditions for training and 11 tests under the same three loading conditions, as listed in Table 1.

The dataset includes acceleration signals that are recorded from both horizontal and vertical directions nearby bearings. The signals were recorded for 0.1 s at 10-s intervals.

### A. Experimental Setup

In order to decompose frequency components from the signals, short-time Fourier transform (STFT) was performed using the entire length of records (0.1 s) as a window length. The Hanning window [21], which also has a 0.1-s, window was applied to eliminate noise. As the sampling rate of the dataset was 25,600 Hz, the size of the frequency band was 1281 with a resolution of 10 Hz.

TABLE 1
OPERATING CONDITIONS OF BEARING DATASETS

| Shaft Rotation Frequency, Maximum Load | 1800 RPM, 4000 N | 1650 RPM, 4200 N | 1500 RPM, 5000 N |
|---|---|---|---|
| Training set | Bearing 1-1 | Bearing 2-1 | Bearing 3-1 |
| | Bearing 1-2 | Bearing 2-2 | Bearing 3-2 |
| Test set | Bearing 1-3 | Bearing 2-3 | Bearing 3-3 |
| | Bearing 1-4 | Bearing 2-4 | |
| | Bearing 1-5 | Bearing 2-5 | |
| | Bearing 1-6 | Bearing 2-6 | |
| | Bearing 1-7 | Bearing 2-7 | |

The obtained power spectra were processed using octave-band filtering, which has a constant absolute band width $m$ of 32 and 16 octave bands in one octave for the following constant percentage band width $n$. As an example, Fig. 3 shows the octave-band filtering of the first record of Bearing 1-1. The graphs on the left show the power spectra after Fourier transform, and the graphs on the right show the processed results. The processed results were converted to a linear scale to compare the graphs. After applying octave-band filtering, the size of the power spectra for each direction was reduced to 88, which is about 6% of the original data size.

The processed power spectra were converted to a two-dimensional matrix, where the number of rows corresponds to the number of the frequency components in the spectra data, and the number of columns corresponds to the number of records. The magnitudes of the matrix were expressed as decibels with a base unit of $10^{1/10}$ $g^2$/Hz. The matrix was fed to an AE of a deep learning model after the normalization process that is described as follows:

$$x'_{i,j} = \frac{x_{i,j} - \mu}{\sigma}, i = 1,2,\ldots,N, j = 1,2,\ldots,M$$

$$\mu = \frac{1}{M'}\sum_{j=1}^{M'} x_{i,j}, \sigma = \sqrt{\frac{1}{M'}\sum_{j=1}^{M'}(x_{i,j}-\mu)^2},$$

$$M' = \lfloor 0.8 \cdot M \rfloor \qquad (6)$$

The element at the $i$th row and $j$th column in the matrix $x_{i,j}$ is normalized by the mean and the standard deviation of the first 80% of the $i$th row elements. This normalization process omits the last 20% of elements because the elements have abnormal values, which shifts the mean and the standard deviation of the row elements.

The neural network model consists of a five-layer AE and a three-layer FFNN. The first three layers of the AE compress the matrix to four features, which is the output of the third layer. The last two layers of the AE, including the third layer, restore the matrix using the features. The FFNN estimates RUL by receiving the features. The depth of the network was minimized while not compromising RUL accuracy too much through trial and error. Figure 4 depicts the structure of the neural networks and the configurations of each network.

All neural layers in the model process their inputs based on the formulae as follows:

$$z^l = W^l \cdot x^l + b^l, x^{l+1} = f(z^l) \qquad (7)$$

where $x^l$ is the input matrix of the $l$th neural layer. The $l$th layer's weight matrix and bias matrix are represented as $W^l$ and $b^l$. The number of rows of the weight matrix $W^l$ and the length of the bias matrix $b^l$ correspond to the number of neurons in the layer. The result of weight sum with the addition of bias $z^l$ is inputted to an activation function $f$, and the output of the function $x^{l+1}$ becomes the next input of a following neural layer. The activation functions in the model were selected according to the input and output range of each neural layer. The formulae of activation functions used in the network are as follows:

$$f(z) = \frac{1}{1 + e^{-z}} \qquad (8)$$

$$f(z) = \frac{1}{e^z - 1}(z < 0), z(z \geq 0) \qquad (9)$$

$$f(z) = \max(0, z) \qquad (10)$$

The first layer of the AE uses a sigmoid function in (8) to change the range of the matrix values from $(-\infty, \infty)$ to $(0,1)$. The last layer of the AE uses an exponential linear unit (ELU) function [22] in (9) because the range of an ELU covers negative values. As most of the normalized outputs of the AE for training ranges (–3, 3), the output of the last layer was multiplied by 3 to express the ranges of the output. The FFNN uses a rectified linear unit (ReLU) function [23] for the last layer since RULs are always positive. The rest of the layers in the model use an ELU because the layers have no constraints on the range of input and output, and an ELU trains the network faster than other activation functions [22].

The weights and biases of the neural networks were optimized by minimizing a cost function $E$ that is based on Olshausen and Field's work [24] to avoid overfitting training data. The function comprises a mean squared error between inputs $x'$ and outputs $\widehat{x'}$, a sparsity regularization term $\Omega_{\text{sparsity}}$, and a L2 regularization term $\Omega_{\text{weights}}$ as follows:

$$E = \frac{1}{N}\sum_{i=1}^{N}(x'_i - \widehat{x'}_i)^2 + \lambda \cdot \Omega_{\text{weights}} + \beta \\ \cdot \Omega_{\text{sparsity}} \qquad (11)$$

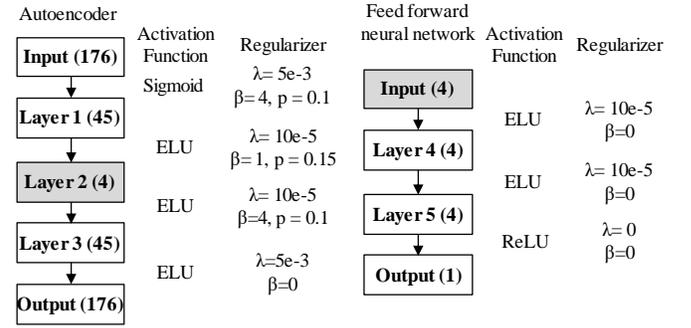

Fig. 4. Neural structures of an autoencoder and a feedforward neural network that consists of the developed deep learning model. Layer 2 in the autoencoder is reused as an input layer for the feedforward layer. The hyperparameter settings that include the type of activation functions and coefficients for regularization terms are listed next to the arrows between the blocks.

The first term in (11) represents the mean squared error between the input $x'$ and output $\widehat{x'}$ of the AE for $N$ observations. The second term $\Omega_{\text{weights}}$ is an L2 regularization term that limits utilizing all inputs to compress and recover inputs because some inputs may not contribute to the network's performance. The term squares all weights that are associated with neurons in the neural network as follows:

$$\Omega_{\text{weights}} = \frac{1}{2}\sum_{l}^{L}\sum_{i}^{N}\sum_{j}^{M}(w^l_{ij})^2 \qquad (12)$$

where $w^l_{ij}$ represents the weight of the $j$th neuron at the $l$th layer for the $i$th input.

A sparsity regularization term $\Omega_{\text{sparsity}}$ in (12) controls the average output values that are generated by a neural layer. The term uses Kullback-Leibler (KL) divergence [25] to measure the discrepancies between the desired average activation value $\rho$ and the calculated average activation value $\hat\rho$ as follows:

$$\Omega_{\text{sparsity}} = \sum_{l=1}^{L} KL(\rho||\hat\rho_l) \\ = \sum_{l=1}^{L} \rho \log\left(\frac{\rho}{\hat\rho_l}\right) + (1-\rho)\log\left(\frac{1-\rho}{1-\hat\rho_l}\right) \qquad (13)$$

$$\hat\rho_l = \frac{1}{M}\sum_{j}^{M}|f(z^l_j)|$$

The average activation value at the $l$th layer $\hat\rho_l$ is obtained by averaging the absolute outputs at the $l$th layer. The output $f(z^l_j)$ in (13) is the output of activation function at the $l$th layer by inputting the $j$th neuron's weighted sum, as shown in (7). The coefficients $\lambda$ and $\beta$ in (11) work as weights that define the importance of each regularization term. The values of the coefficients vary by the type of network and the depth of the neural layers.

The AE and the FFNN in the model were trained in a supervised manner, using Keras for model training. The inputs and outputs of the AE were normalized power spectra after octave-band filtering. The spectra data were randomly selected from both training and test sets because the amount of training data in Table 1 was limited for the AE to learn generalized features. The test sets that were used for the training were not full run-to-failure data. They were given for RUL predictions at the end of the data. The first 50% of the combined data were used for training, and remaining 50% were used for validation.

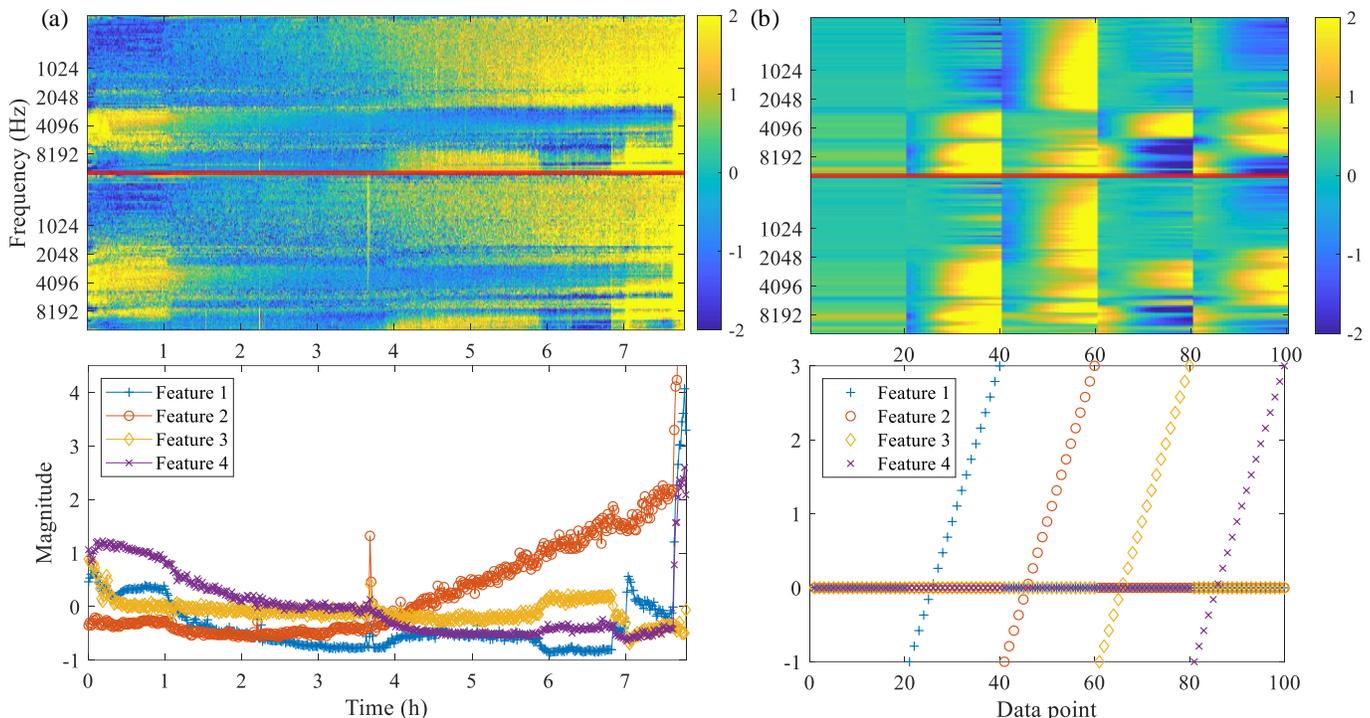

Fig. 5. High-level feature visualization. (a): response of the high-level features of the trained autoencoder due to the changes in the magnitudes of inputs of the autoencoder. (b): response of the outputs of the autoencoder due to the changes in the high-level feature values.

The FFNN was trained by feeding it with labeled data that consist of the outputs of the trained AE's bottleneck layer, Layer 2, in Fig. 4, and the RUL of rotating machinery that is associated with the outputs. The RULs were calculated backward from the end data point of the run-to-failure data, and the maximum RUL was truncated. The maximum RUL for each data was set to the RUL of the first data point, where an anomalous behavior in high-level features is detected. The anomaly for each data was detected with the help of a Bayesian change-point detection method developed by Lavielle [26]. The method provides change-points where the mean of the high-level feature changed the most. Since some training data have more than two change-points, the maximum RULs were determined using the last change-points of the data.

The truncated RUL was rescaled to (0, 1) while training the FFNN. The rescaling process helps the FFNN to maintain the same scale of weight distributions over layers, which stabilizes weight optimization. The 70% of the training sets in Table 1 are used to train the FFNN after shuffling, and the remaining 30% are used as validation data. The RUL estimation accuracy of the model was evaluated using all the test data in Table 1.

### B. Neural Network Inspection

The AE and FFNN in the model were inspected after training. The characteristics of the high-level features of the AE were inferred by visualizing them. The graphs in Fig. 5 give insights about how the AE interprets inputs and composes the features. The graph on the top left shows the spectrogram of Bearing 1-1 data in Table 1 over time. The spectrogram is constructed by stacking power spectra of horizontal accelerations on top of power spectra from vertical accelerations. The x-axis of the graph represents the operation time of the bearing, and the magnitude of each frequency component at a certain time is represented using a color bar that scales the magnitude to (-2, 2). The graph on the bottom left shows the response of high-level features as the inputs of the AE change. The data in the graph are the outputs of neurons from the AE bottleneck when the spectra of the graph above are received as inputs.

One method to characterize the high-level features is reactive monitoring. The method couples the magnitude changes of the features and the changes in the spectrogram in certain regions. For example, during the first operation hour, features 1 and 4 respond to the magnitude changes in the mid-range frequency band that ranges from 2773 Hz to 6888 Hz for the horizontal axis and 2048 Hz to 8192 Hz for the vertical axis. Also, these features respond to magnitudes of frequencies over 8192 Hz for both directions after 4 hours operation. On the other hand, feature 1 value monotonically increases as the RUL of a bearing approaches 0. This behavior can be inherited by the trends of magnitudes at the low-frequency band that ranges from 0 Hz to 2048 Hz for both directions. This inference of the features' behavior is intuitive. However, the inference should be validated by many observerations over samples.

Direct feature injection is a complementary feature analysis method to the reactive monitoring. As the AE can recover the input of the model with high-level features, the composition of individual high-level features can be explained by injecting controlled high-level features into the AE's decoder. The graphs on the right in Fig. 5 show the changes in the output of the AE as the value of the feature changes. The graph on the bottom shows the inputted high-level features, and the graph on the top shows the output of the AE. The feature values increased one at a time from -1 to 3, leaving the remaining feature values at 0 because most of the feature values were distributed within the ranges over samples. As shown by the former analysis, features 1 and 4 contribute to the changes in the magnitude of mid-range frequency components for both directions. When both feature values increase, the magnitudes of the mid-range frequency band (2773 Hz – 6888 Hz) also increase.

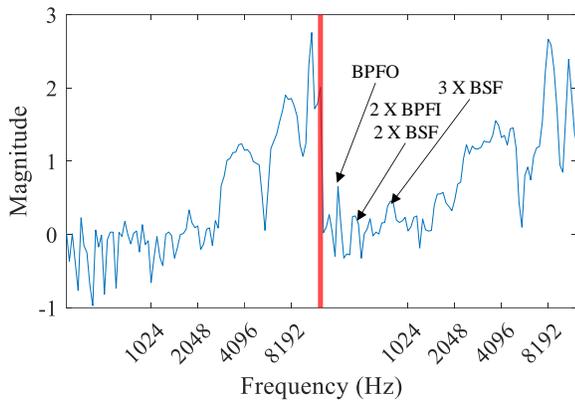

Fig. 6. The magnitudes of frequency components at data point 30 in Fig. 5. The first 88 frequency components are related to horizontal accelerations, and the rest of the components are related to vertical accelerations.

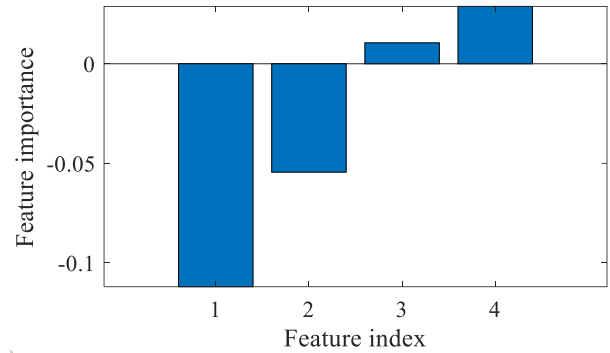

Fig. 7. The importance of four high-level features of an autoencoder when a feedforward neural network estimates the remaining useful lives of bearings.

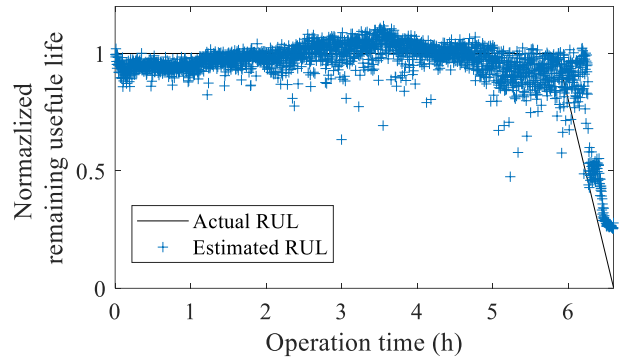

Fig. 8. Comparison of actual and estimated remaining useful life using Bearing 1-3 data.

Compared to the former method, this method can segregate the effects of each feature when both features are responsive to the same frequency bands. Feature 1 has a positive correlation with the magnitudes of frequency bands over 6888 Hz. On the other hand, feature 4 has a negative correlation with the same magnitudes. The composition of feature 1 and feature 4 can be determined through the analysis,. This method also helps humans assign a physical meaning for each feature. Since each feature highlights certain frequency bands, relating the characteristic frequencies of bearings to the highlighted frequency bands may help researchers understand the physics behind the observations. For example, some frequency components that correlate with feature 1 coincide with the characteristic frequencies of a bearing that were analyzed by Wang's work [27]. The graph in Fig. 6 shows the magnitudes of frequency components at data point 30 in the top right graph in Fig. 5. Three peaks in the graph are related to a ball pass frequency outer race (BPFO), the second harmonic of a ball pass frequency inner race (BPFI) or a ball spin frequency (BSF), and the third harmonic of a BSF. Likewise, the local maxima of other features were related to the characteristic frequencies, as listed in Table 2.

The local maxima located at frequency bands over 3 kHz may indicate another physical meaning. The location of the maxima can be related to the resonance of bearing components [17]. The frequency components that have central frequencies of 3839, 7678, and 11340 Hz in a horizontal direction and 3371, 5199, 5921 Hz in a vertical direction responded to the features.

TABLE 2
OBSERVED CHARACTERISTIC FREQUENCIES FROM HIGH-LEVEL FEATURES

|  | Feature 1 | Feature 2 | Feature 3 | Feature 4 |
|---|---|---|---|---|
| Characteristic frequency | 1 X BPFO | 2 X BPFO | 1 X BPFO | 2 X FTF |
|  | 2 X BPFI |  |  | 2 X BPFI |
|  | 2 X BSF |  |  | 2 X BSF |
|  | 3 X BSF |  |  |  |

The analyzed features of the AE are used as the inputs of the FFNN. Feature importance was measured to evaluate the contribution of each feature to the result of the FFNN. Figure 7 shows the measured feature importance of the FFNN using the connection weight method. The FFNN relies on feature 1 the most and relies on feature 3 the least to estimate RULs. Feature 1 and 2's importance values are negative because both features have a negative correlation with RULs. This result coincides with those of the high-level feature analysis. The changes in feature 1 are related to the magnitudes of four characteristic frequencies of a bearing. Feature 2's monotonically increasing trend is regarded as a good characteristic for a health indicator. On the other hand, feature 3's trend is the least correlated with the RULs. Also, feature 3 may not deliver unique information for prognosis. Many frequencies that are responsive to feature 3's changes are also responsive to features 1 and 4's changes.

### C. Experimental Results

The developed model outputs normalized RULs when inputting modified spectral data. Figure 8 shows the developed model's estimated RULs using Bearing 1-3 data as an example. Since the anomalous behavior of the high-level feature was detected at 6.84 h, the actual RUL in the graph decreases after the anomaly point. The estimated RULs in the graph fluctuate because the inputs of the model have measurement noise.

The performance of the developed model was evaluated with three other machine learning models: a least-squares linear regression (Linear) model, a support vector regression (SVR) model, and a one-dimensional convolutional neural network (1-

D CNN). The Linear model was trained using Scikit-learn [28] as machine learning tools. The Linear model receives the high-level features of the AE of the developed model to estimate RULs. The Linear model's performance provides insights into the nonlinearity between the high-level features and RULs. If the Linear model's performance is the same as that of the developed model, the features have a linear relationship with RULs rather than a nonlinear relationship.

The SVR model and the 1-D CNN were developed by Yang et al. [29]. The SVR model represents the performance of conventional prognostics that use time and frequency domain statistical features as inputs. The 1-D CNN provides the performance of a typical deep learning model that does not process raw data.

The performance of the developed model and the other models was evaluated using root mean squared errors (RMSEs) and relative errors (REs) [30] as metrics that have been applied for evaluating the performance of the SVR and the 1-D CNN. Since Yang's work [29] provides the RMSEs and RAs of Bearing 1-1, Bearing 1-2, Bearing 1-3, and Bearing 1-6, the performance of the Linear model and the developed model calculated the RMSEs and RAs for the same data. All errors were calculated from the last point of a truncated true RUL for each data. The performance of the developed model was evaluated by averaging five RMSEs and RAs while retraining the model because the weights and biases of the model converge to different sub-optimal points due to lack of training data.

The performance of the four machine learning models is compared in Table 3.

| | | Developed Model | Linear | SVR [28] | 1D-CNN [28] |
|---|---|---|---|---|---|
| Bearing 1-1 | RMSE (s) | 554.2 | 533.6 | N/A | 377.4 |
| | RE | 0.58 | 0.51 | N/A | 0.96 |
| Bearing 1-2 | RMSE (s) | 90.39 | 74.2 | N/A | 21.3 |
| | RE | -0.02 | 0.45 | N/A | 0.94 |
| Bearing 1-3 | RMSE (s) | 1821.4 | 2093.5 | 102.8 | 43.0 |
| | RE | 0.17 | -0.61 | 0.34 | 0.77 |
| Bearing 1-6 | RMSE (s) | 4741.7 | 4333.5 | 7275.5 | 720.3 |
| | RE | -2.75 | -3.37 | -1.51 | 0.78 |

The RMSEs and REs of Bearing 1-1 and Bearing 1-2 show the performance of the models on training data, and the metrics for other data represent the performance of the models on test data. The RMSEs and REs of the SVR on training data were left empty because they were not provided.

The performance difference between Linear and the developed model was no significant because the high-level features of the AE have a linear relationship with the RULs. In general, these models showed better RMSE than the RMSE of SVR, but 1D-CNN outperformed all three models. However, 1D-CNN was highly curated to degradation patterns for the test data. Unlike the developed model and Linear, 1D-CNN uses two models to generalize two different degradation patterns, and the models were trained by single bearing data from operation under one condition (1800 RPM, 4000 N). The performance of 1D-CNN and SVR can degrade if they are trained using all bearing data under different operating conditions.

The RE of the developed model was inferior to the REs of other models for two reasons. The RE performance is highly dependent on discrepancies between true RULs and estimated RULs that are close to 0. As shown in the estimated RULs in Fig. 8, the developed model tends to provide progressive RULs. The offset between true RULs and the estimated RULs resulted in high REs. In addition, the developed model did not apply additional methods such as a Kalman filter to correct measurement errors and system errors. On the other hand, 1D-CNN uses a weight method that handles the variances in the estimated RULs.

The characteristics of the developed model's high-level features are revealed by comparing the Linear and the SVR model's performance. The conventional statistical features have a limitation in generalizing the raw data. Even a simple linear regression model was able to outperform the conventional SVR model.

## IV. DISCUSSION

This paper introduced octave-band filtering as a knowledge-based dimensionality reduction method and hypothesized that octave-band filtering contributed to reducing the number of layers of the trained AE. This section discusses the necessity of octave-band filtering in training the developed model. The validity of octave-band filtering as a replacement for the function of low-level layers is also evaluated in the session by training two deep learning models.

In order to validate the need for octave-band filtering, the same developed structure was trained by inputting spectral data without octave-band filtering. The trained model is prone to underfit data because it requires additional hidden layers to generate low-level features from the data. The underfitting was found when the high-level features of the model were inspected. Figure 9 visualizes the high-level features in the same manner as Fig. 5 visualizes the features of the original AE. Although the developed model can learn four different trends in inputs, feature 4 did not learn meaningful characteristics of the inputs. This result proves that octave-band filtering affects the training of deep learning models.

As a second deep learning model, a convolutional autoencoder (CAE) was trained to verify the replaceability of the low-level features's funtion of a deep learning model. Since the low-level convolutional layers of CNN-based diagnostics can function as filters [8, 9], the second model replaced the functionality of octave-band filtering of the developed model by adding convolutional layers to the model. The model adds two convolutional layers at the beginning and the end of the AE of the original model.

The feature analysis results of the CAE in Fig. 10 coincide with the results of the original AE. For convenience, the order of the features in Fig. 10 was modified to align with the orders of the features in Fig. 5 that show similar behaviors. Many portions of the behaviors and the compositions of the CAE's high-level features overlapped with those of the AE. For example, feature 1 in the AE changes the magnitudes of frequency bands at around 4096 Hz and 8192 Hz in horizontal accelerations. Feature 1 in the CAE also has correlations with the magnitudes at the same frequency bands, which indicates that the function of the convolutional layers in the CAE resembles the output of the octave-band filtering.

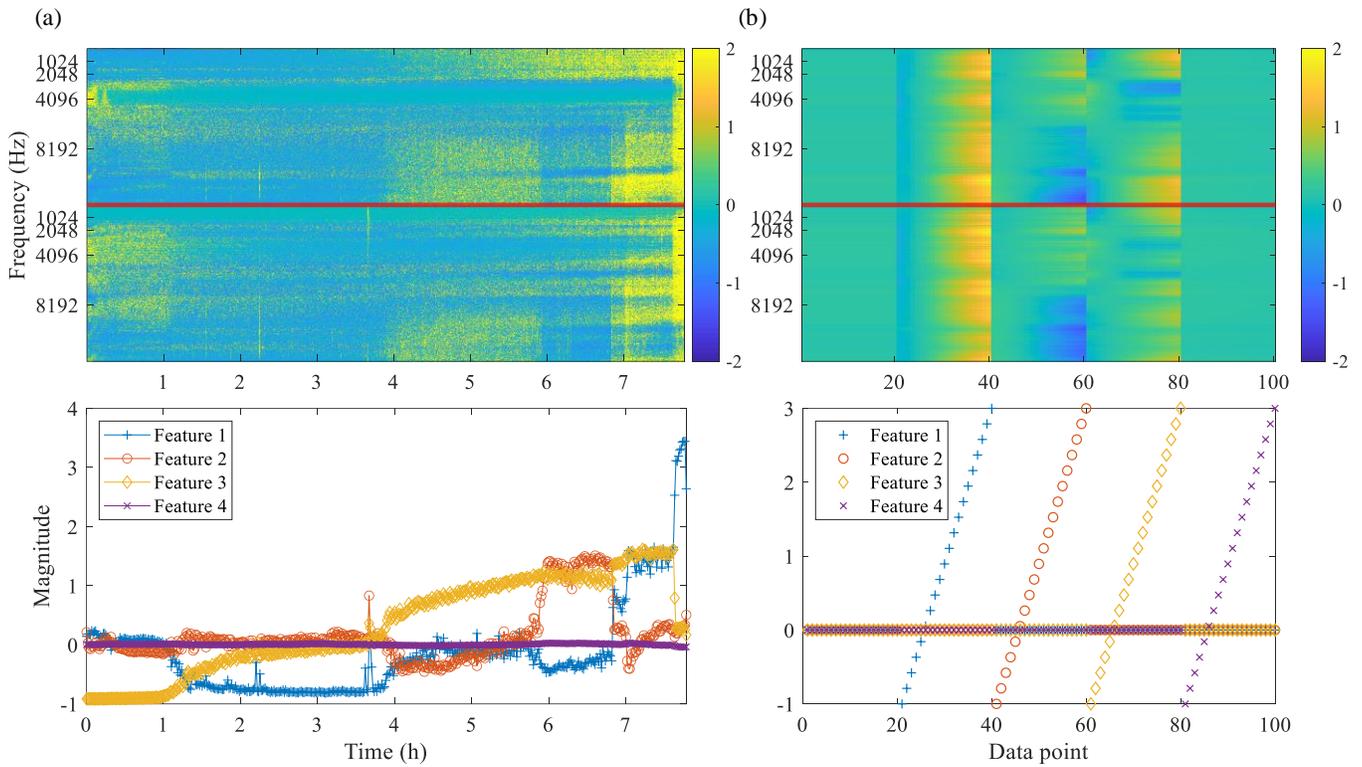

Fig. 9. An exemplary feature visualization result. (a): response of the high-level features of an autoencoder due to the changes in the magnitudes of inputs of the autoencoder. (b): response of the outputs of the autoencoder due to the changes in the high-level feature values.

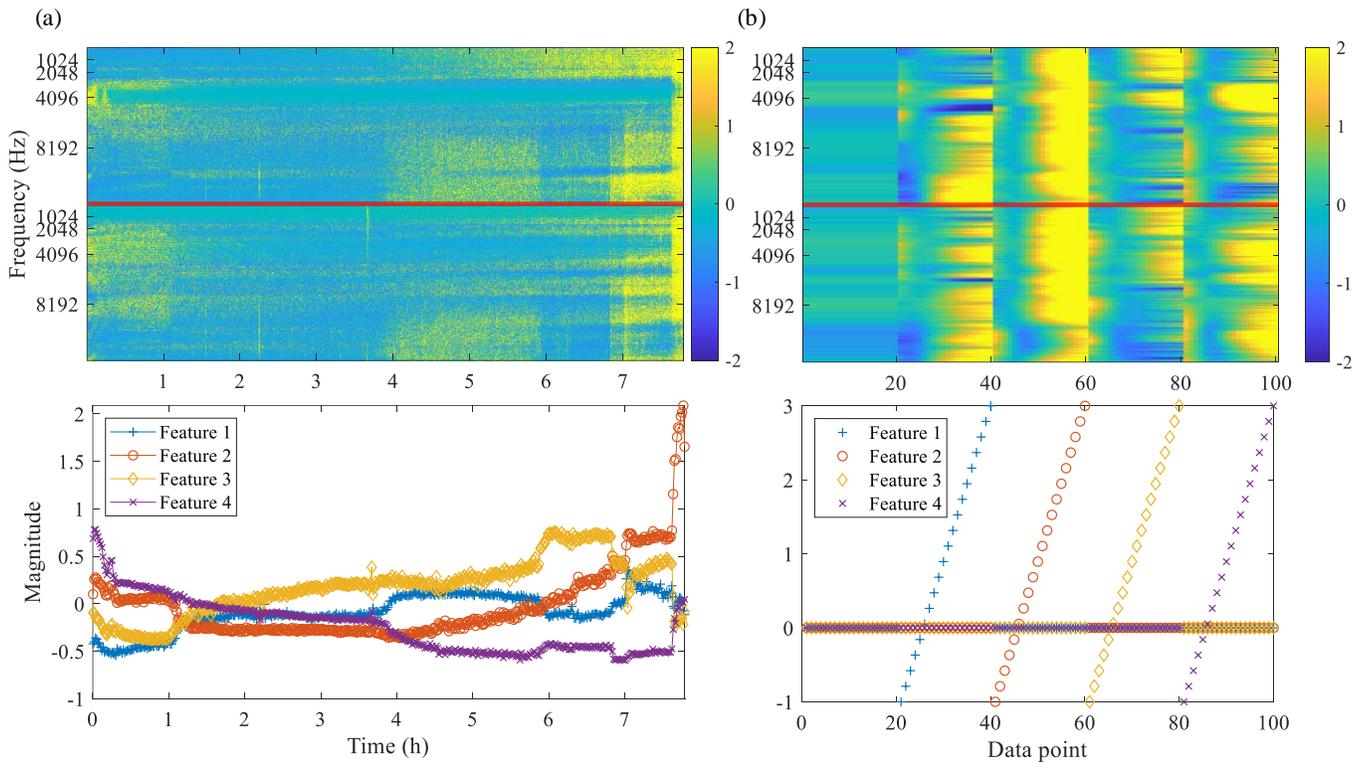

Fig. 10. An exemplary feature visualization result. (a): response of the high-level features of the complex autoencoder due to the changes in the magnitudes of inputs of the autoencoder. (b): response of the outputs of the autoencoder due to the changes in the high-level feature values.

Compared to the original AE, the CAE can characterize the high-resolution inputs since it uses raw spectral data. However, the CAE has a drawback in analyzing high-level features since the added convolutional layers increase the complexity to understand the features. The complex nonlinear relationship among high-level features interferes with predicting the individual feature behaviors. Overall, all four features in the bottom left graph in Fig. 10 are constantly changing over time and collaboratively constitue the spectral data. As a result, the introduced reactive monitoring and direct feature injection methods may not apply to deep learning models that have too many layers. The experiment demonstrates the need for the depth reduction method to improve the interpretability of the model.

## V. CONCLUSIONS

Interpretability of a deep learning model is one of the concerns in applying deep learning for prognostics. Although many deep learning-based prognostic models have been developed, only a few models are able to inspect portions of the model's behavior. This paper developed an explainable deep learning-based prognostic model for rotating machinery with the help of domain knowledge.

The developed model consists of an autoencoder (AE) and a feedforward neural network (FFNN). The AE extracts high-level features from inputs, and the FFNN estimates the remaining useful life (RUL) of rotating machinery using the features. The model receives processed power spectra of vibration signals through modified octave-band filtering. Since the filtering process reduces the size of the AE's input, the depth of the AE can be reduced to a level where humans can understand the model's behavior.

A case study showed the feasibility of explaining the behavior by training the model for bearing datasets. The physical meaning of high-level features in the trained model was explained by monitoring the changes of the feature values as the values of the inputs changed. In addition, the composition of each feature was determined by manipulating the feature values one at a time and then observing the changes in the outputs of the AE. The features responded to specific frequencies related to the characteristic frequencies of a bearing and the resonance frequencies of bearing components.

The case study also inspected the FFNN in the developed model by evaluating the contribution of each input to the output to understand the basis of the output. The contributions were measured by comparing the connection weights associated with the inputs.

The performance of the developed model was evealuted by comparing it to four other machine learning algorithms using the same datasets. The developed model's RUL estimation was more accurate than the conventional support vector regression model. However, the developed model underperformed compared to a deep learning model that uses raw data as inputs. This result explains the trade-off between the performance and interpretability of a deep learning model.

The efficacy of using octave-band filtering as a low-level convolutional layer was demonstrated through additional experiments. Without octave-band filtering, the developed model underfitted data, and one of the model's high-level features did not learn a meaningful pattern from the inputs. Another experiment was conducted to train a convolutional autoencoder (CAE) to substitute the function of octave-band filtering using convolutional neural layers. The feature analysis results of the CAE demonstrated that the convolutional layers could work as filters similar to octave band filters. The experiment also exhibited the trade-off between interpretability and the ability to capture patterns in inputs.

The developed model can be applied to any system that utilizes vibration analysis through frequency decomposition. However, the model may not work on a complex system that needs to be trained using a much deeper AE than the one used in the case study. As the depth of a neural network increases, the nonlinearity between input and output of the network may increase, which hampers evaluation of the contributions of each feature to the final outputs of the network. Therefore, another method should be developed to explain a complex neural network modeland will be a future study.

## REFERENCES

[1] Z. Li, J. Li, Y. Wang, and K. Wang. "A deep learning approach for anomaly detection based on SAE and LSTM in mechanical equipment," *Int. J. Adv. Manuf. Technol.* vol. 103, pp. 499–510, Jul. 2019.

[2] W. Lu, Y. Li, Y. Cheng, D. Meng, B. Liang and P. Zhou, "Early Fault Detection Approach With Deep Architectures," *IEEE Trans. Instrumentation and Measurement*, vol. 67, no. 7, pp. 1679-1689, July 2018.

[3] M. Ma, C. Sun and X. Chen, "Deep Coupling Autoencoder for Fault Diagnosis With Multimodal Sensory Data," *IEEE Trans. Ind. Informatics*, vol. 14, no. 3, pp. 1137-1145, Mar. 2018.

[4] B. Yang, R. Liu, and E. Zio, "Remaining Useful Life Prediction Based on a Double-Convolutional Neural Network Architecture," *IEEE Trans. Ind. Electron.*, vol. 66, no. 12, pp. 9521-9530, Dec. 2019.

[5] M. Zhao, M. Kang, B. Tang, and M. Pecht, "Multiple Wavelet Coefficients Fusion in Deep Residual Networks for Fault Diagnosis," *IEEE Trans. Ind. Electron.*, vol. 66, no. 6, pp. 4696-4706, Jun. 2019.

[6] Saeki, M., J. Ogata, M. Murakawa, and T. Ogawa, "Visual explanation of neural network based rotating machinery anomaly detection system," *IEEE Int. Conf. Prognostics and Health Manage.*,2019, pp. 1-4.

[7] J. Grezmak, J. Zhang, P. Wang, K. Loparo, and R. Gao, "Interpretable Convolutional Neural Network through Layer-wise Relevance Propagation for Machine Fault Diagnosis," *IEEE Sensors Journal*, Dec. 2019.

[8] F. Jia, Y. Lei, N. Lu, and S. Xing, "Deep normalized convolutional neural network for imbalanced fault classification of machinery and its understanding via visualization." *Mech. Syst. Signal Process.*, vol 110, pp. 349-367, Sep. 2018.

[9] Y. Lei, F. Jia, J. Lin, S. Xing, and S. X. Ding, "An Intelligent Fault Diagnosis Method Using Unsupervised Feature Learning Towards Mechanical Big Data," *IEEE Trans. Ind. Electron.*, vol. 63, no. 5, pp. 3137-3147, May 2016.

[10] S. Dreiseitl, and L. Ohno-Machado. "Logistic regression and artificial neural network classification models: a methodology review." *Journal of biomedical inform.* Vol. 35. no. 5-6, pp. 352-359, Oct. 2002.

[11] S. Bach, A. Binder, G. Montavon, F. Klauschen, K. Müller, and W. Samek. "On pixel-wise explanations for non-linear classifier decisions by layer-wise relevance propagation," *PloS One* vol. no. 7, 2015.

[12] M. Ribeiro, S. Singh, and C. Guestrin. "Why should I trust you?: Explaining the predictions of any classifier," *Proc. ACM SIGKDD Int. Conf. Knowledge Discovery and Data Mining*. pp. 1135-1144, 2016.

[13] S. Lundberg, and S. Lee. "A unified approach to interpreting model predictions," *Adv. Neural Inform. Processing Sys.* Pp. 4765-4774, 2017.

[14] Y. LeCun, Y. Bengio, and G. Hinton. "Deep learning." *Nature* vol. 521 pp. 436-444, May. 2015.

[15] D. Ballard, "Modular learning in neural networks", *Proc. National Conf. Artificial Intelligence*, pp. 279-284, 1987.

[16] *Specification for Octave-Band and Fractional-Octave-Band Analog and Digital Filters*, ANSI Standard S1.11-2004.

[17] B. Graney, and S. Ken. "Rolling element bearing analysis," *Materials Eval.*vol. 70, no.1, pp. 78, Jan. 2012.

[18] J. Olden, M. Joy, and R. Death. "An accurate comparison of methods for quantifying variable importance in artificial neural networks using simulated data," *Ecol. Model.* vol. 178 no. 3-4, pp. 389-397, Nov. 2004.

[19] J. Olden, and D. Jackson, 2002b. "Illuminating the "black box": a randomization approach for understanding variable contributions in artificial neural networks," *Ecol. Model.* vol. 154, no. 1-2, pp. 135–150, Aub. 2002.

[20] P. Nectoux, R. Gouriveau, K. Medjaher, E. Ramasso, B. Chebel-Morello, N. Zerhouni, and C. Varnier "Pronostia: An experimental platform for bearings accelerated degradation tests," *IEEE Int. Conf. Prognostics and Health Manage*, 2012, pp. 1–8.

[21] A. Oppenheim, R. Schafer, and J. Buck, *Discrete-Time Signal Processing*, Upper Saddle River, NJ: Prentice Hall, 1999.

[22] D. Clevert, T. Unterthiner, and S. Hochreiter. "Fast and accurate deep network learning by exponential linear units (elus)," 2015, *arXiv:1511.07289*.

[23] V. Nair, and G. Hinton. "Rectified linear units improve restricted Boltzmann machines," *Int. Conf. Machine Learning*, 2010, pp. 807-814.

[24] B. Olshausen, and D. Field. "Sparse Coding with an Overcomplete Basis Set: A Strategy Employed by V1." *Vision Research*, vol. 37, no. 23, pp.3311–3325, Dec. 1997.

[25] S. Kullback, and R. Leibler. "On information and sufficiency," *The annals Math. Stat.* vol. 22 no. 1 pp. 79-86, Mar. 1951.

[26] M. Lavielle, "Using penalized contrasts for the change-point problem," *Sig. Processing* vol. 85 no. 8 pp. 1501-1510, Aug. 2015.

[27] T. Wang, "Bearing life prediction based on vibration signals: A case study and lessons learned," *IEEE Int. Conf. Prognostics and Health Manage.,*2012, pp. 1-4.

[28] F. Pedregosa, G. Varoquaux, A. Gramfort, V. Michel, B. Thirion, O. Grisel, M. Blondel, P. Prettenhofer, R. Weiss, V. Dubourg, J. Vanderplas. "Scikit-learn: Machine learning in Python," *Journal of machine learning research.* vol. 12, pp. 2825-30, Oct. 2011

[29] B Yang, R Liu, E Zio. "Remaining useful life prediction based on a double-convolutional neural network architecture." *IEEE Trans. Ind. Electron.,* vol. 66, no. 12, pp. 9521-9530, Jul. 2019.

[30] A. Saxena, J. Celaya, B. Saha, S. Saha, and K. Goebel, "Metrics for offline evaluation of prognostic performance," Int. J. Prognostics Health Manage., vol. 1, no. 1, pp. 4–23, 2010.